\documentclass[letterpaper, 10 pt, conference]{ieeeconf}  

\IEEEoverridecommandlockouts                              
\usepackage{graphicx}
\usepackage{amsmath}
\usepackage{amssymb}
\usepackage{siunitx}
\usepackage{times} 
\usepackage{balance}
\usepackage{subcaption}
\usepackage[font=small]{caption}     % chữ nhỏ
\usepackage[maxfloats=256]{morefloats}
\makeatletter
\let\NAT@parse\undefined
\makeatother
\usepackage[bookmarks=false, linkcolor=blue, urlcolor=blue, citecolor=blue]{hyperref} 
\usepackage{fancyhdr}
\hypersetup{
     colorlinks=true,
     linkcolor=red,
     filecolor=magenta,      
     urlcolor=blue,
     pdfstartview={FitH},
     citecolor =blue
     }
\title{\LARGE \bf
Soft–Rigid Hybrid Gripper with Inflatable Silicone Pockets for\\Tunable Frictional Grasping           
}

\author{Hoang Hiep Ly, Cong-Nhat Nguyen*, Doan-Quang Tran, Quoc-Khanh Dang, Ngoc Duy Tran, Thi Thoa Mac, \\ Anh Nguyen, Xuan-Thuan Nguyen, Tung D. Ta}

\begin{document}

\maketitle
\thispagestyle{empty}
\pagestyle{empty}

%%%%%%%%%%%%%%%%%%%%%%%%%%%%%%%%%%%%%%%%%%%%%%%%%%%%%%%%%%%%%%%%%%%%%%%%%%%%%%%%
\begin{abstract}
Grasping objects with diverse mechanical properties, such as heavy, slippery, or fragile items, remains a significant challenge in robotics. Conventional rigid grippers typically rely on increasing the normal forces to secure an object; however, this can cause damage to fragile objects due to excessive force. To address this limitation, we propose a soft-rigid hybrid gripper finger that combines rigid structural shells with soft, inflatable silicone pockets, which could be integrated into a conventional gripper. The hybrid gripper can actively modulate its surface friction by varying the internal air pressure of the silicone pockets, enabling the gripper to securely grasp objects without increasing the gripping force. This is demonstrated by fundamental experimental results, in which an increase in internal pressure leads to a proportional increase in the effective coefficient of friction. The gripping experiments also show that the integrated gripper can stably lift heavy and slippery objects or fragile, deformable objects, such as eggs, tofu, fruits, and paper cups, with minimal damage by increasing friction rather than applying high force. 
\end{abstract}

% ==================== FIGURE 1 (ĐẦU TRANG PHẢI) ====================

% =================================================================

%%%%%%%%%%%%%%%%%%%%%%%%%%%%%%%%%%%%%%%%%%%%%%%%%%%%%%%%%%%%%%%%%%%%%%%%%%%%%%%%
\section{Introduction}
% Introduction text goes here
In the field of robotic, grippers are key components, serving as the end-effector. It plays a crucial role in the interaction between the robot and its environment, particularly in robotic manipulation such as pick-and-place tasks within logistics, bio-medical, and service fields \cite{Shintake2018}, \cite{Hughes2016}. The gripper efficiency is dependent not only on the gripping force (normal force), but also on the contact conditions between the gripper and the objects, such as surface friction, object geometry, and mechanical properties. In practical applications, robots are often required to handle a variety of objects with diverse physical properties within a single operation. This is challenging when grasping heavy, slippery items (which may cause sliding or dropping) or fragile objects (where excessive gripping force may cause deformation or surface damage) \cite{Marco2013}, \cite{Achilli2020}. Thus, designing a gripper that ensures grasping stability while maintaining the safety of the object remains a challenge in robotic manipulation \cite{Hughes2016}.

\begin{figure}[h!]
    \centering
    \includegraphics[width=\linewidth]{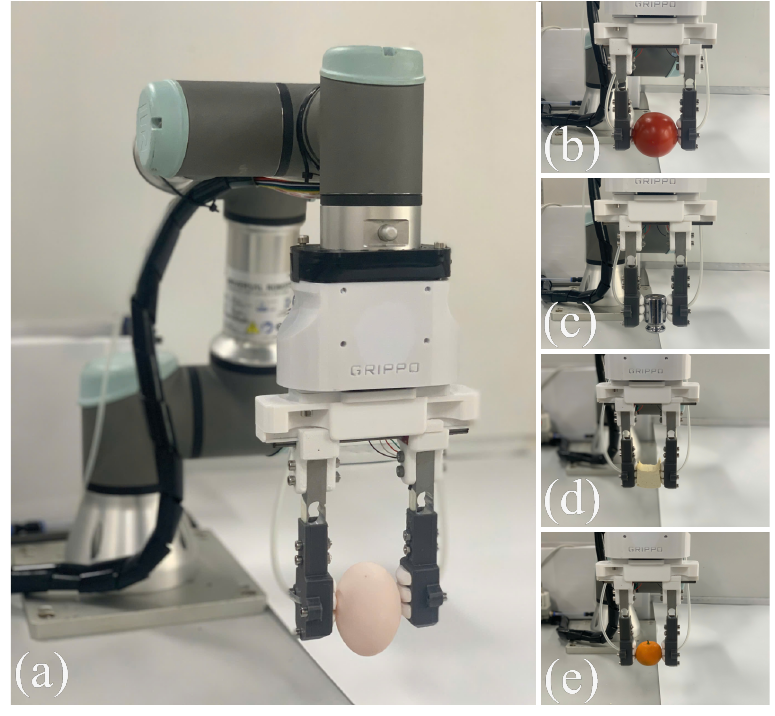}
    \caption{Hybrid robotic gripper for adaptive grasping using inflatable silicone pockets to grasp heavy and slippery objects, deformable objects, and various types of objects (a) egg, (b) tomato, (c) stainless steel weight, (d) tofu, (e) orange. }
    \label{fig:concept_overview}
    \vspace{-0.7cm}
\end{figure}

Rigid grippers are currently the most common type used in the industry \cite{Reddy2013}. These typically feature parallel-jaw or angular-jaw configurations, with the gripper components fabricated from metallic or hardened materials~\cite{Liu2020, Nishimura2018, Do2023, Ko2020, Wei2023}. The operating principle of rigid grippers relies on generating normal force at the contact surface to generate sufficient friction to prevent slippage. The advantages of this type include simple structure, durability, high precision, and the ability to exert large gripping forces. However, the gripper rigidity limits their adaptability to diverse object shapes and often results in small contact areas. Due to low surface friction, when rigid grippers handle objects with smooth surfaces, they must increase the gripping force to enhance effective friction, leading to persistent slippage risks or damage to fragile objects resulting from the high contact stress \cite{Guo2017}. Additionally, positioning errors of the robot can lead to eccentric grasping, causing slippage or surface scratching \cite{Guo2017}. In contact mechanics, the increase of the friction coefficient is a key method for improving anti-slip performance. This method helps reduce the required gripping force, which minimizes potential damage to the objects in the cases mentioned above. A common solution is applying rubber pads to the rigid gripper fingers. However, these pads often have high stiffness and small effective contact areas, leading to poor conformal grasping on complex or curved surfaces, which is disadvantageous for fragile or scratch-sensitive objects.

Soft grippers have seen rapid development recently due to their flexible shape-morphing capabilities, allowing them to conform to objects and distribute contact pressure more effectively, thereby increasing adaptability and safety\cite{Ham2018},\cite{Chen2021}. Soft grippers can be classified into three groups \cite{Białek2024}. The first group consists of soft grippers based on external force interactions, using geometric deformation to conform to the object, leading to increasing contact area and stability \cite{Park2018}, \cite{Zhou2017}, \cite{Yang2017}. These are often actuated employing pneumatic, tendon-driven, or electromechanical mechanisms to control their soft structure and flexibility. The second group involves stiffness-controllable soft grippers, where the stiffness of the contact interface can be modulated during manipulation using mechanisms like jamming \cite{Amend2012}, \cite{Tsugami2017}. In the gripper type, the end-effector is soft to adapt to the object shape and increase in stiffness for stable lifting. The third group focuses on adhesion-controlled grippers, often using suction or electroadhesion to increase holding force without increasing gripping force \cite{Daler2013}, \cite{Yue2022}. This type of soft gripper is suitable for convex, flat, or deformable objects. 

Despite their high safety and adaptability, soft grippers are often limited by low gripping force, poor positioning accuracy, slow operation speeds, and instability in high-noise environments. This is a drawback for industrial integration where high payloads are required alongside the handling of slippery or fragile items. In order to create a gripper that has the high payload of rigid grippers with the high adaptability of soft grippers, hybrid grippers with rigid-soft structures have been proposed \cite{Białek2024}. Some designs use a rigid frame with joints made of soft materials; by varying the stiffness at these joints, the gripper can handle diverse objects \cite{Tran2025}. Another approach uses a rigid frame with soft fingers to increase contact area and reduce stress concentration. Thus, these grippers can manipulate a wide range of objects, from heavy items requiring high force to slippery or fragile ones, without causing damage. Some hybrid grippers employ Magnetorheological (MR) fluids in the soft fingers, but this material is sensitive to magnetic fields \cite{Białek2024}. Others use pneumatic power to create balloon-shaped fingers, which are unaffected by magnetic fields but often require a large physical size to remain stable \cite{Cao2023} \cite{Choi2006}. Thus, creating a gripper that combines stable gripping with high shape adaptability remains a challenge.

In this paper, we propose a hybrid gripper using a soft-rigid finger that enables friction modulation during grasping via inflatable silicone pockets. With this structure, surface stiffness and contact area can be actively adjusted by modulating the air supply, allowing for the adjustment of the effective friction between the finger and the object. This finger design aims to enhance anti-slip stability and grasping efficiency for heavy, slippery, and fragile objects without the need for increased gripping force. The soft contact interface, comprising multiple silicone pockets, ensures a wider, more uniform distribution of the contact area while maintaining a compact gripper finger size. The remaining paper presents the theory of the correlation between pneumatic pressure and effective friction; subsequently, the construction of the gripper finger and the integrated gripper system are described; then, experimental evaluations of the friction modulation mechanism and the effectiveness of the proposed gripper in handling diverse objects are presented and analyzed.

%%%%%%%%%%%%%%%%%%%%%%%%%%%%%%%%%%%%%%%%%%%%%%%%%%%%%%%%%%%%%%%%%%%%%%%%%%%%%%%%
\section{Methods and Materials}

\subsection{Hybrid gripper finger in the context of contact theory}
\begin{figure}[tp]
    \centering
    \includegraphics[width=.9\linewidth]{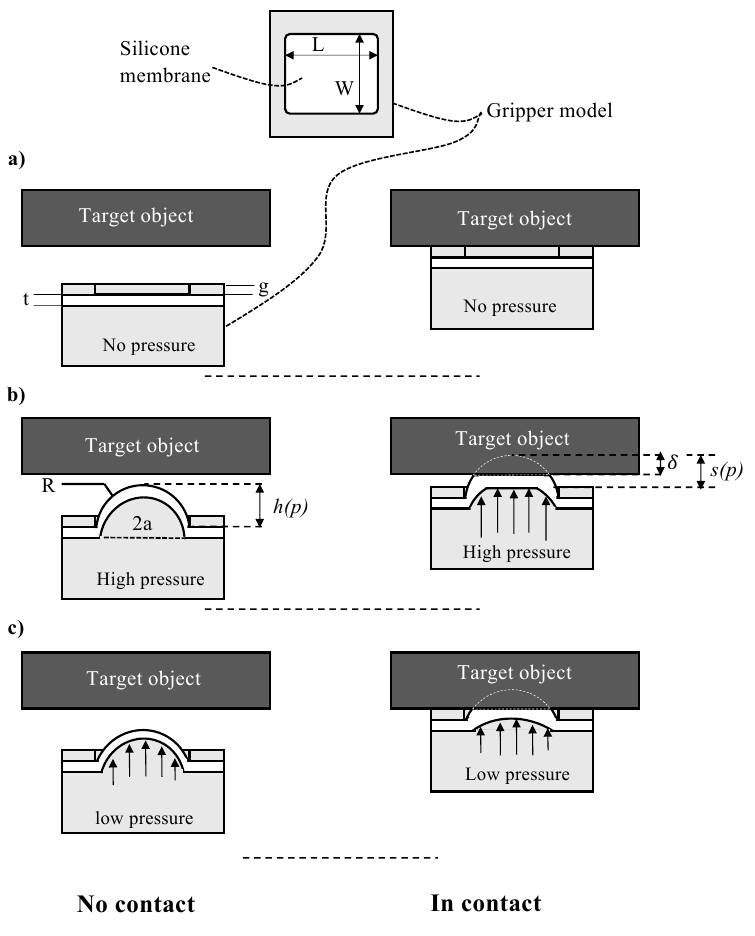}
    \caption{Modeling the contact mechanisms between a hybrid gripper finger and a target object. a) Contact in no pressure condition b) Contact in high pressure condition c) Contact in low pressure condition.}
    \vspace{-1.5em}
    \label{fig:contact_mechanism}
\end{figure}

In this study, the contact phenomenon between a hybrid gripper finger and the surface of an object is modeled in a simplified manner, as shown in Fig.~\ref{fig:contact_mechanism}. Assume the gripper is a rigid box whose opening is a rectangle with dimensions of $W\times L$ and $(W<L)$, sealed by a soft silicone membrane of thickness $t$; the rim of the box is recessed by a gap $g$ below the plane of the metal target surface, as in Fig.~\ref{fig:contact_mechanism}. When an internal pressure $p$ is applied, the silicone membrane bulges into a spherical cap with apex deflection of $h(p)$. According to the bulge test~\cite{Xiang2005}, the relationship between $p$ and $h$ is presented as follows,
\begin{equation}
    p = 2\frac{\sigma_0 t}{a^2}h + \frac{4}{3}\frac{Et}{(1-\nu^2)a^4}h^3
\end{equation}
where $\sigma_0, a, E, \text{and } \nu$ denote the in-plane equibiaxial residual stress, the effective half-span ($0.5 W$), the Young’s modulus of the silicone membrane, and the Poisson’s ratio, respectively. For small $h$, $h \approx k_h p$ where $k_h = \frac{a^2}{2\sigma_0 t}$, therefore,
\begin{equation}
    h(p) = \min(k_h p, h_{\max})
\end{equation}
in which $k_h$ is the proportionality coefficient between bulge height and pressure, $h_{\max}$ is the limiting (critical) bulge height of the silicone membrane. Suppose the bulged portion forms a spherical cap; then the effective radius of curvature is presented as follows,
\begin{equation}
    R(p) = \frac{a^2+h(p)^2}{2h(p)}
\end{equation}
when $h(p) > g$, the protrusion beyond the rim can be computed as follows,
\begin{equation}
    s(p) = h(p) - g
\end{equation}
Because the pressurization tensions the membrane, an effective contact stiffness arises and is expressed as follows,
\begin{equation}
    E^*(p) = E_0(1+\eta p)
\end{equation}
where $E_0$ is the effective Young’s modulus at $p=0$ and $\eta$ is the proportionality factor describing how the effective stiffness scales with $p$. For a given normal load $N$, the contact area $A(N,p)$ and indentation $\delta(N,p)$ are given by the following equations,
\begin{equation}
    A(N,p) = \pi a_c^2
\end{equation}
\begin{equation}
    \delta(N,p) = \left[ \frac{3N}{4E^*(p)\sqrt{R(p)}} \right]^{2/3}
\end{equation}
where $a_c = \left( \frac{3NR(p)}{4E^*(p)} \right)^{1/3}$ is the contact radius of the bulged silicone against the object surface. Using Archard's Elastic Model of Friction~\cite{JFArchard1957} the friction force ($F_f$) is presented as follows,
\begin{equation}
    F_f(N,p) = \tau_s A(N,p) = \tau_s \pi \left( \frac{3R(p)}{4E^*(p)} \right)^{2/3} N^{2/3}
    \label{eq:friction_force}
\end{equation}
where $\tau_s$ is the interfacial shear stress for the silicone–object contact. Consequently, the friction coefficient of the silicone $\mu_s(N,p)$ against the object is
\begin{equation}
    \mu_s(N,p) = \tau_s \pi \left( \frac{3R(p)}{4E^*(p)} \right)^{2/3} N^{-1/3}
    \label{eq:friction_coef}
\end{equation}

With the above model, the silicone’s contact with the target plane can be categorized into three cases: no inner pressure (Fig.~\ref{fig:contact_mechanism}a), high inner pressure (Fig.~\ref{fig:contact_mechanism}b), and low inner pressure (Fig.~\ref{fig:contact_mechanism}c). In the absence of inner pressure—or when the pressure is too low so that $h(p) < g$ the friction is that between the rigid rim of the box and the target object, leading to a small friction force. When the pressure is increased so that $h(p) > g$ but remains insufficiently large, the appearance of a normal load $N$ pushes the bulged silicone downward; then $s(p) \le \delta$, and part of the box rim contacts the object, which reduces the gripper–object friction coefficient (Fig. ~\ref{fig:contact_mechanism}c). In contrast, when the inner pressure is high enough, for the same normal load $N$ the silicone fully contacts the object ($s(p) \ge \delta$), increasing the friction with the coefficient given by (\ref{eq:friction_coef}). In this case, the friction coefficient is clearly tunable via the inner pressure $p$. This is the theoretical basis for our proposed gripper, whose friction coefficient can be adjusted by controlling the inner pressure.

\subsection{Hybrid Gripper Finger Design}
\begin{figure}[bp]
    \centering
    % NOTE: You must provide an image file named 'fig3.jpg' for this figure.
    \includegraphics[width=\columnwidth]{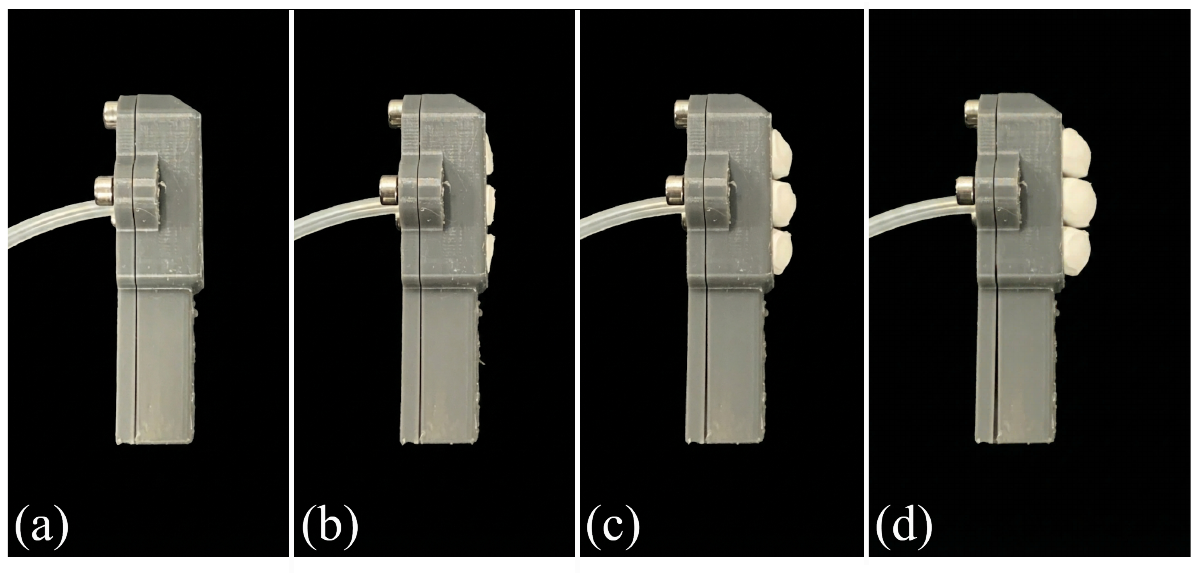}
    \caption{Inflation stages of the silicone air pocket at different pressure levels. (a) Total deflated, (b) Slightly inflated, (c) Inflated, (d) Fully inflated.}
    \label{fig:inflation_stages}
\end{figure}

The proposed gripper finger comprises two components: a rigid outer shell that provides structural support during grasping, and a soft silicone air pocket housed inside the shell that inflates to increase friction when gripping an object. When air pressure increases in the air pockets, they expand outwards. At higher pressure levels, the pockets inflate to a greater degree, expanding until they reach the elastic limit of the silicone. These various stages of inflation are illustrated in Fig.~\ref{fig:inflation_stages}.

% ==================== Figure 2 ====================

% ===============================================================

% ==================== NEW Figure 3 ====================

% ===============================================================

\subsubsection{Outer Shell}
\begin{figure}[tp]
    \centering
    \includegraphics[width=0.9\linewidth]{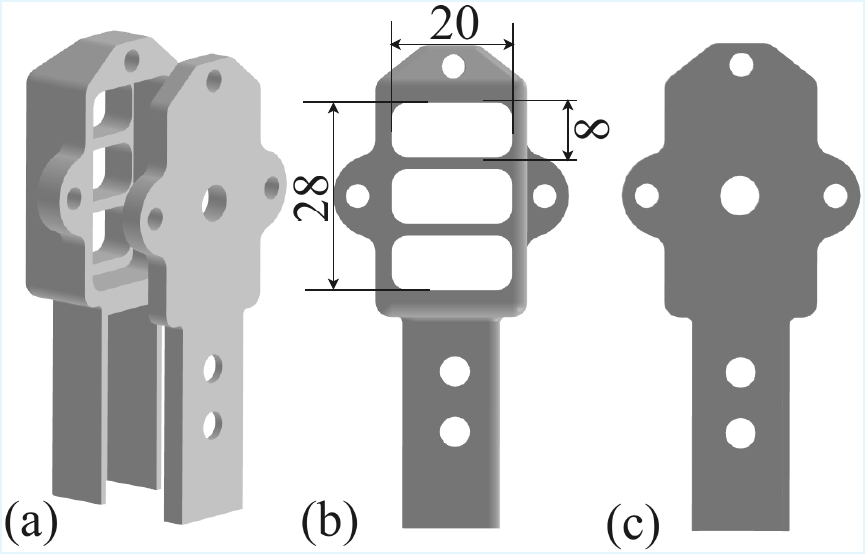}
    \caption{Design model of the outer shell. (a) Assembled view. (b) Front part with three grooves. (c) Back part with load-cell mount. Dimensions are in \SI{}{\milli\meter}.}
    \label{fig:outer_shell}
    \vspace{-0.3cm}
\end{figure}

Three grooves (cross-section of \SI{20}{\milli\meter}$\times$\SI{8}{\milli\meter}) were added to the front surface of each finger, allowing the internal silicone pocket to expand and induce bending, as shown in Fig.~\ref{fig:outer_shell}. For ease of assembly, the finger was divided into two parts. The front part contains a hollow cavity to accommodate the silicone air pocket; the grooves enable outward expansion upon inflation. The back part serves as a cover and includes a circular port for the air tube and mounting features for the load cells. The outer shell was fabricated by 3D printing using PETG.

% ==================== Figure 4 (was 3) ====================
% ===============================================================

\subsubsection{Silicone Air Pocket}
The air pocket is the soft component placed inside the outer shell. It was molded from silicone and designed to fit the internal geometry of the shell: a rectangular base with three rounded bulges on the front side, corresponding to the grooves of the outer shell. This design ensures a tight fit and allows the silicone to expand outward when pressurized. Compressed air is supplied through a silicone tube (inner diameter \SI{2}{\milli\meter}, outer diameter \SI{4}{\milli\meter}) inserted on the back side of the pocket.

\begin{figure}[htbp]
    \centering
    \includegraphics[width=\linewidth]{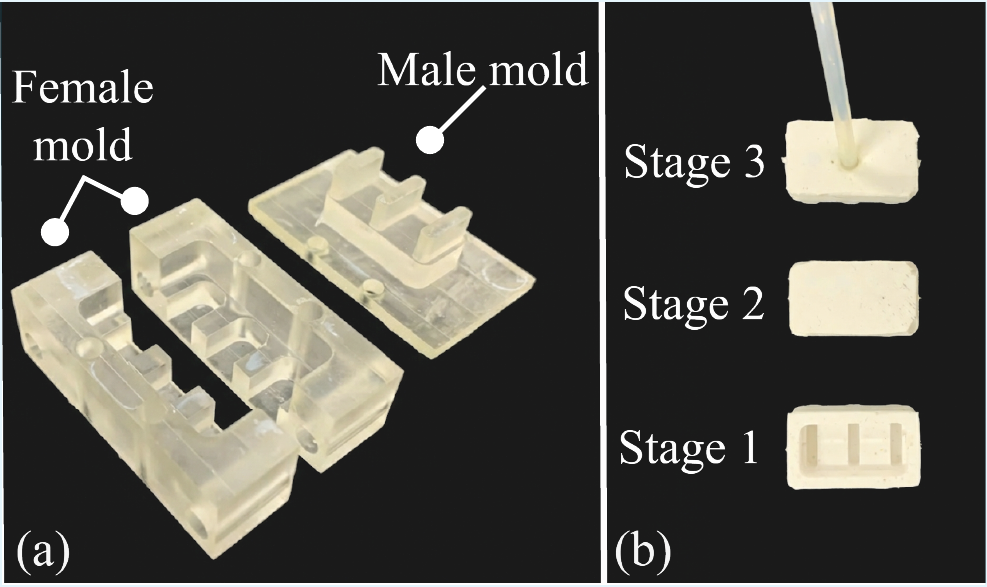}
    \caption{Fabrication of the silicone air pocket. (a) Female mold (Split in half for easy assembly and disassembly during molding) and Male mold. (b) The three stages of fabrication. }
    \label{fig:silicone_mold}
    \vspace{-0.5cm}
\end{figure}

The air pocket was fabricated by molding. The male and female molds were fabricated by a 3D printer (Formlabs) as illustrated in Fig.~\ref{fig:silicone_mold}(a). The silicone (Shin Etsu KE-1416) was poured into the female mold, after which the male mold was pressed on top and fixed using screws through pre-designed holes. After curing, this process produced a silicone pocket with one open face. In the second step, another mold was used to seal the open surface, creating a closed air chamber. Finally, a small hole was punched in the back to insert the air tube. These three fabrication stages are illustrated in Fig.~\ref{fig:silicone_mold}(b).

% ==================== Figure 5 (was 4) ====================
% ===============================================================

\subsection{Hybrid Gripper Prototype}
The silicone air pocket was inserted into the outer shell and sealed by fastening the two shell parts together with screws. The air pressure generated from a pump was provided to the silicone pocket using air tubes, as shown in Fig.~\ref{fig:inflation_stages}. Two fingers were then mounted on one side of two $\SI{1}{\kilo\gram}$-load cell sensors, while the opposite side of each load cell was fixed to a support frame consisting of a linear guide and a sliding rail. A stepper motor (NEMA 17) was employed to actuate the gripper jaws, enabling both opening and closing actions. The full gripper assembly is shown in Fig. \ref{fig:concept_overview} a). To perform the experimental evaluations, the hybrid gripper was integrated with a 6-DoF (UR3) robot. High-level motion planning was autonomously executed via the ROS MoveIt framework, while a microcontroller independently managed the low-level hardware (stepper motor, pressure valve, and load cells). Both subsystems were seamlessly synchronized through a ROS master node to enable a fully autonomous grasping pipeline. 
For pneumatic actuation, compressed air was supplied by an air compressor and regulated using an electro-pneumatic regulator (SMC, ITV1030). The valve input voltage ranged from \SIrange{0}{5}{\volt}, corresponding to an output pressure of \SIrange{0}{500}{\kilo\pascal}. The regulator was controlled by the microcontroller, assisted by a digital-analog converter (DAC) module (MCP4725) for voltage control. 

% ==================== Figure 6 (was 5) ====================
\begin{figure}[htbp]
    \centering
    \includegraphics[width=\linewidth]{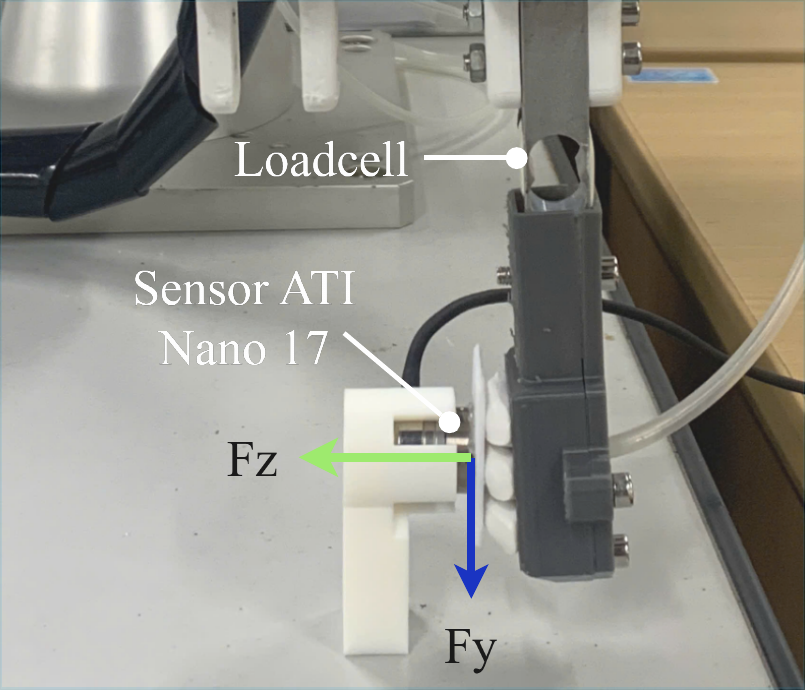}
    \caption{  Fundamental Experiment set up, the finger will move upward along Y-Axis.\textcolor{red}{}}
    \label{fig:new_figure_3}
    \vspace{-0.5cm}
\end{figure}
% ===============================================================

%%%%%%%%%%%%%%%%%%%%%%%%%%%%%%%%%%%%%%%%%%%%%%%%%%%%%%%%%%%%%%%%%%%%%%%%%%%%%%%%
\section{Experiments}

\subsection{Fundamental Experiment}
Fundamentally, an increase in applied pressure causes integrated air pockets to expand from the surface, resulting in a corresponding increase in frictional force, as mentioned in the previous section. To validate the hypothesis, this experiment utilizes a 6-axis force/torque sensor (ATI Nano17) and a data acquisition device (National Instruments). The sensor is mounted to a table with its Z-axis oriented perpendicularly to the finger to measure the normal force. At the same time, its Y-axis is aligned parallel to the finger's direction of motion, as shown in Fig.~\ref{fig:new_figure_3}. The procedure is conducted by pushing the finger to the sensor's surface to apply a normal force ($F_{z}$), and then sliding the finger upwards across the surface. During this motion, the sensor records the tangential force ($F_{y}$) as the friction force, while $F_{z}$ represents the normal force.  Different materials (paper, plastic, and rubber) were used to cover the force sensor to determine the coefficient of friction between the gripper finger and these materials, which is calculated by the following equation,
\begin{equation}
\mu_s =\frac{\left|F_{y}\right|}{\left|F_{z}\right|}
\end{equation}
Data is acquired from the moment the finger begins its upward slide until it is no longer in contact with the surface. All values of $F_{y}$  and $F_{z}$ will be converted to their absolute values.  To ensure the reliability of the results and address potential sensor noise, the normal force $F_{z}$ was maintained at a constant level around $3\text{ N}$ across all experimental trials. This specific magnitude was selected based on preliminary tests, which indicated that $3\text{ N}$ is sufficient to ensure that all silicone bulges establish complete and uniform contact with the sensor surface, even at zero or low inflation pressures. Consequently, this provides a consistent baseline for measuring friction modulation. Furthermore, the coefficient of friction $\mu_s$ was determined by calculating the average value of the ratio $\frac{\left|F_{y}\right|}{\left|F_{z}\right|}$ over the entire duration of the upward sliding motion, from the onset of movement until the point of contact separation. 
\subsection{Grasping Objects Experiments}
\subsubsection{Grasping Heavy and Slippery Objects Experiments}

The objective of this experiment is to evaluate the efficacy of the hybrid gripping mechanism in grasping heavy and slippery objects, a task that poses a significant challenge for conventional grippers. The underlying principle is that as the pressure within the integrated air pockets is increased, the silicone surface expands through apertures in the rigid shell. This action enlarges the effective contact area between the finger and the object, which in turn generates a higher frictional force and significantly enhances grasp stability.

The experiment utilized two steel weights (slippery payloads) of \SI{200}{\gram} and \SI{500}{\gram}. The evaluation was conducted using an automated pick-and-place sequence executed by the UR3 robot framework, as shown in Fig.~\ref{fig:concept_overview}(c). The gripper's fingers first approached the payload to ensure perpendicular contact with its parallel surfaces, then applied a compressive force (adjusting the inner pressure) before lifting. To systematically characterize the impact of friction modulation, the system was programmed to autonomously perform multiple grasping cycles with the internal pneumatic pressure incrementally increased in steps: \SIlist{0;25;50;75;100;125}{\kilo\pascal}.

For each (normal force, pressure) setpoint, the following force-thresholding strategy was autonomously executed: (i) the pneumatic system regulated the silicone pockets to the target pressure; (ii) the fingers closed until the integrated load cells detected the pre-set target normal force; (iii) once this threshold was reached, the motor held its position, and the UR3 arm performed the transport motion. The experiment investigated three constant normal force levels for each payload: \SIlist{3;3.5;4}{\newton} for the \SI{200}{\gram} object, and \SIlist{8;8.5;9}{\newton} for the \SI{500}{\gram} object. This protocol ensures that the normal force remains consistent across different pressure levels, effectively isolating the friction enhancement effect from the mechanical gripping force.

At each setpoint, 10 grasp trials were performed to determine the success rate. A trial was defined as successful if the gripper could stably grasp and move the object from its initial position to the target location without dropping it.

To lift the object, the total static friction force ($F_s$) generated by the gripper fingers must be greater than or equal to the object's weight ($W_{o}$).
\begin{equation}
    F_s \geq W_{o} = mg
\end{equation}
where $F_s$ is the total static friction force and $W_{o}$ is the gravitational force on the object.

To generate this friction, the gripper must apply a normal force ($F_z$) perpendicular to the object's surfaces. This force is measured directly by the integrated load cells. The static friction force is proportional to the normal force via the static coefficient of friction ($\mu_s$):
\begin{equation}
    F_{s} = \mu_s F_{z}
\end{equation}
where $F_z$ is the normal gripping force and $\mu_s$ is the static coefficient of friction between the fingertip and the payload.

From this, the minimum required normal force to prevent the object from slipping can be determined:
\begin{equation}
    F_z \geq \frac{mg}{\mu_s}
\end{equation}

\subsubsection{Grasping Deformable Objects Experiments}
\begin{figure}[htbp]
    \centering
    \vspace{-0.3cm}
    % NOTE: You must provide an image file named 'fig7_new.jpg' for this figure.
    \includegraphics[width=\columnwidth]{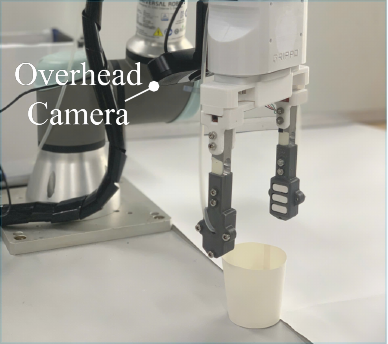}
    \caption{Grasping deformable objects experiment setup with an overhead camera.}
    \label{fig:deformable_setup}
    \vspace{-0.3cm}
\end{figure}

In this experiment, paper cups were used as deformable test objects, with payloads of \SI{100}{\gram} and \SI{200}{\gram} (steel weight) placed inside to simulate loading conditions. Utilizing the UR3 robotic arm and the established force-thresholding protocol, the system was programmed to execute an automated pick-and-place trajectory. The objective was to determine the optimal grasping condition that balances normal force and the provided pressure to the soft pocket, to minimize the cup deformation. To isolate the effects, the experiment was conducted by incrementally increasing either the gripping force $N$, or the inflation pressure $P$ while the control system maintained the other parameter constant. By leveraging the force-feedback from the integrated load cells, we ensured that the gripper stopped closing immediately upon reaching the target force, preventing any unnecessary crushing of the cup. For quantitative evaluation, a camera was mounted above the gripper to capture an image of the cup's rim after each grasping trial. The experiment was conducted 10 times for each payload. The experimental setup is illustrated in Fig.~\ref{fig:deformable_setup}.

The degree of deformation was quantified by the Roundness Ratio\textbf{ ($R$)}, which compares the shortest (minor) diameter - $D_{\min}$  and longest (major) diameter - $D_{\max}$ of the cup's rim post-grasp as follows,
\begin{equation}
    R = \frac{D_{\min}}{D_{\max}}
    \label{equa_roundness}
\end{equation}

$R$ value ranges from 0 to 1. $R$ value approaching 1 indicates that the minor and major diameters are nearly equal, signifying that the paper cup has maintained a nearly perfect circular shape with minimal deformation. Conversely, a smaller $R$ value (approaching 0) indicates a larger discrepancy between the diameters, signifying greater deformation.

\subsubsection{Grasping Different Objects Experiments}
\begin{figure}[htbp]
    \centering
    % NOTE: You must provide an image file named 'fig7_new.jpg' for this figure.
    \includegraphics[width=\columnwidth]{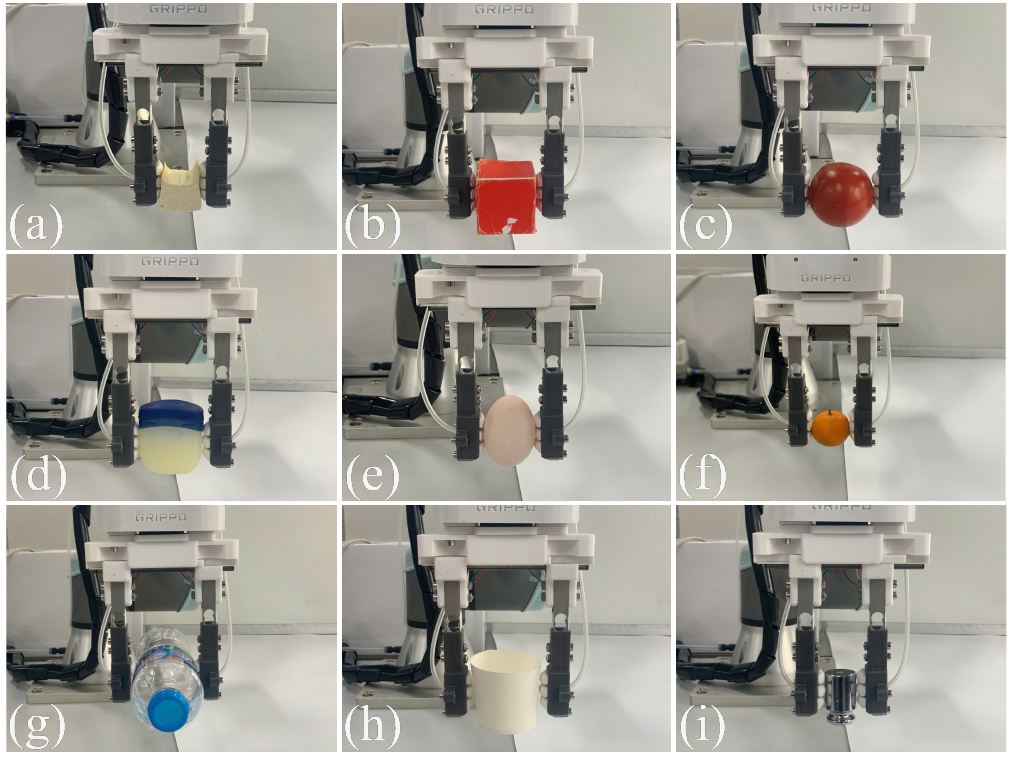}
    \caption{Grasping experiments with various objects. (a) Tofu, (b) Box, (c) Tomato, (d) Vaseline jar, (e) Egg, (f) Orange, (g) Water bottle--(h) Paper cup and (i) Weight from previous experiments. }
    \label{fig:versatility_setup}
    \vspace{-0.5cm}
\end{figure}

In this experiment, a variety of objects with different shapes and surface properties were tested, including tofu, a box, a tomato, a vaseline jar, an egg, an orange, and a plastic bottle, as shown in Fig.~\ref{fig:versatility_setup}. The purpose was to demonstrate the versatility of the gripper in handling a wide range of objects. For each object, an initial gripping force is applied to ensure moderate contact without causing damage. When the internal pressure is at 0 kPa, the object can hardly be lifted. Each object was grasped repeatedly under gradually increasing pressure levels, allowing us to evaluate the gripper’s adaptability and performance across diverse scenarios.

% ==================== NEW Figure 7 ====================
% ===============================================================

%%%%%%%%%%%%%%%%%%%%%%%%%%%%%%%%%%%%%%%%%%%%%%%%%%%%%%%%%%%%%%%%%%%%%%%%%%%%%%%%
\section{Experiment Results}

\subsection{ Fundamental Experiment Results}
\begin{figure}[htbp]
    \centering
    \includegraphics[width=\columnwidth]{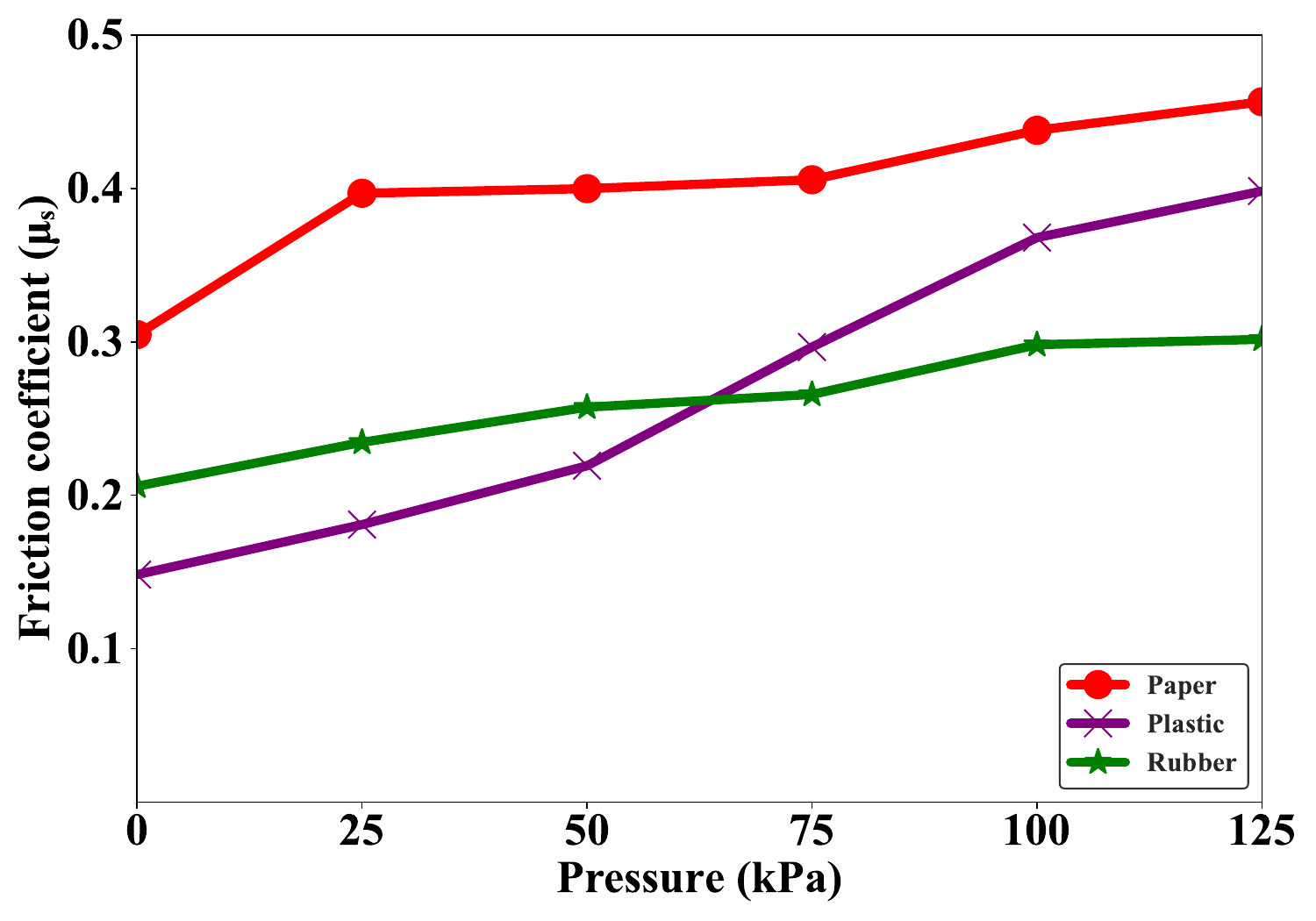}
    \caption{Relationship between the average coefficient of friction ($\mu_s$) and internal pneumatic pressure for different materials (paper, plastic, and rubber). All data points were recorded at a constant normal force of $F_{z} = 3$~N. }
    \label{fig:friction_vs_pressure}
    \vspace{-0.3cm}
\end{figure}

Fig. \ref{fig:friction_vs_pressure} illustrates the average coefficient of friction ($\mu_s$) as a function of increasing applied pressure for various materials, conducted with a constant normal force $F_{z} = 3\text{ N}$. The results demonstrate a clear positive correlation: an increment in internal pressure consistently leads to a significant increase in the effective coefficient of friction across all tested materials (paper, plastic, and rubber). 

% ==================== Figure 8 (was 7) ====================
% ===============================================================

The results support the hypothesis that higher internal pressures cause the silicone pockets to expand and stiffen, thereby increasing the effective contact area and enhancing the frictional interaction. Friction modulation via pneumatic pressure, rather than increased gripping force, enables stable grasping of heavy or slippery objects. This ensures secure handling while protecting fragile items from mechanical damage. 

\subsection{Gripping Objects Experiment Results}

\subsubsection{Grasping Heavy and Slippery Objects Experiments Results}
\begin{figure}[htbp]
    \centering
    \includegraphics[width=\columnwidth]{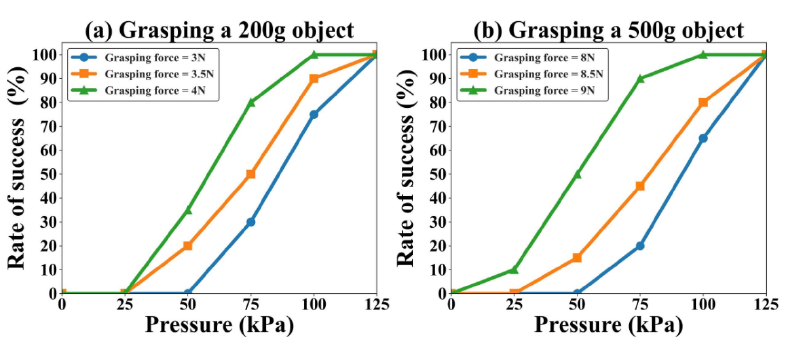}
    \caption{Grasping success rate as a function of pneumatic pressure for different normal force levels. (a) Results of grasping a 200g object (b) Results of grasping a 500g object }
    \label{fig:success_rate_plot}
    \vspace{-0.5cm}
\end{figure}

The grasping experiments for \SI{200}{\gram} and \SI{500}{\gram} steel masses were conducted using a UR3 robotic arm and repeated 10 times for each condition. The grasping success rates for different normal forces ($N$) and internal air pressures ($P$) are summarized in Fig.~\ref{fig:success_rate_plot} (a) and (b), respectively. The normal force thresholds \SIlist{3;3.5;4}{\newton} for the \SI{200}{\gram} payload and \SIlist{8;8.5;9}{\newton} for the \SI{500}{\gram} payload were specifically chosen to represent moderate gripping conditions that avoid excessive mechanical stress on the objects. A strong positive correlation is evident across all trials: for a constant normal force maintained by the robot's control system, the success rate consistently increases as the pneumatic pressure rises. At \SI{0}{\kilo\pascal}, where the gripper effectively functions as a conventional fully rigid gripper, the success rate was 0\%, underscoring the critical role of the pneumatic actuation in transitioning to a hybrid state to enhance grasp stability.

These results provide empirical evidence to validate the initial hypothesis. By holding $N$ constant through the force-thresholding protocol, the observed transition from failed or slipping grasps to stable ones can be attributed solely to the variation of the actuation pressure $P$. This demonstrates that the effective coefficient of friction ($\mu_s$) is a controllable function of pressure. The physical mechanism is likely that the expanded part of the silicone pocket increases the real contact area and creates micro-scale mechanical interlocking with the object's surface, thereby enhancing the friction coefficient.

\begin{figure}[htbp]
    \centering
    \includegraphics[width=\columnwidth]{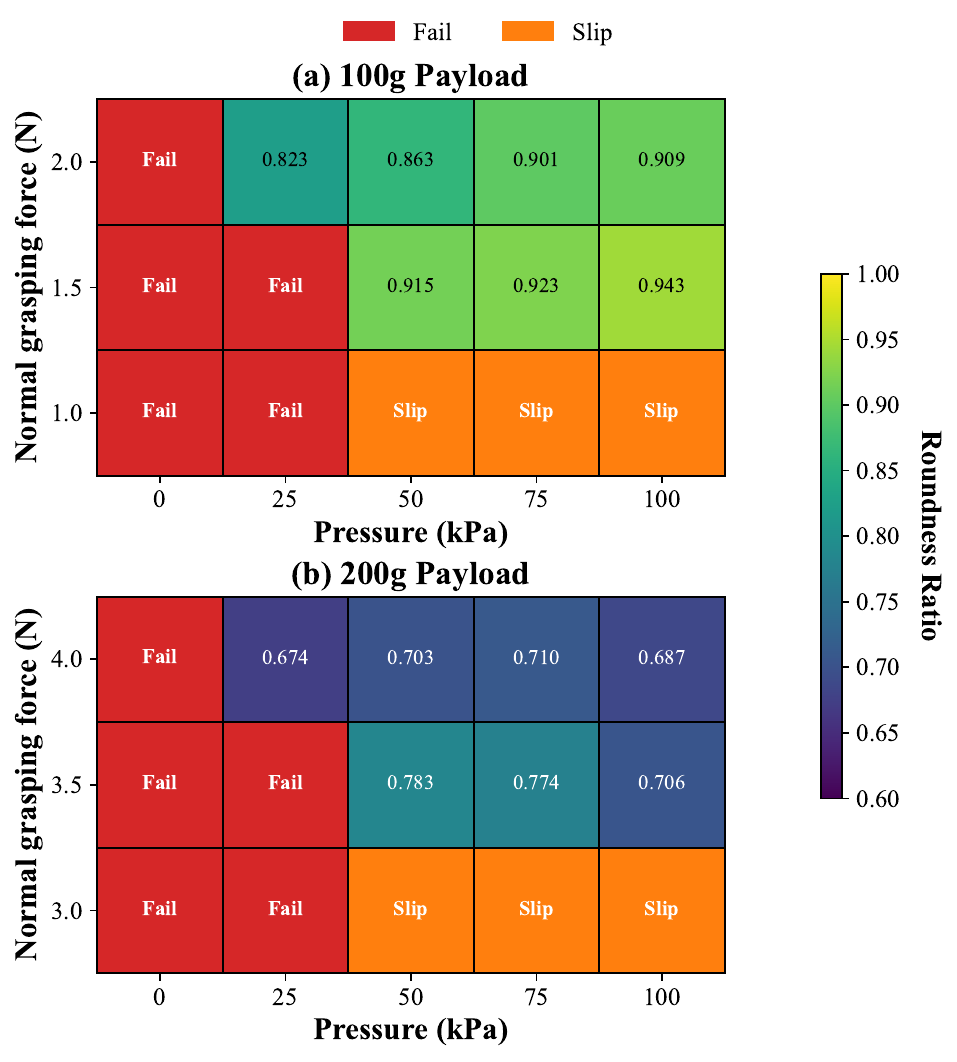}
    \caption{Analysis of deformable object grasping. The map identifies critical failure (red) and slippage (orange) regions, while the roundness ratio when successfully grasping the cup rim is depicted by yellow (maximum roundness)  to dark purple (minimum roundness). (a) Results of grasping a 100g payload (b) Results of grasping a 200g payload.}
    \label{fig:deformable_results}
    \vspace{-0.7cm}
\end{figure}

\subsubsection{Grasping Deformable Objects Experiments Results}
\begin{figure}[htbp]
    \centering
    \includegraphics[width=\columnwidth]{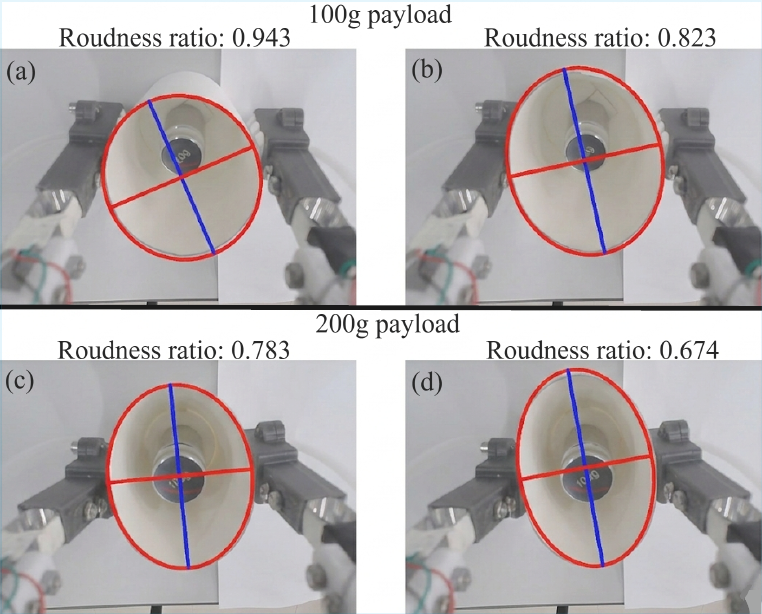}
    \caption{Visualization of the minimum and maximum deformation states of the paper cup. (a) Minimal deformation with a \SI{100}{\gram} payload. (b) Maximal deformation with a \SI{100}{\gram} payload. (c) Minimal deformation with a \SI{200}{\gram} payload. (d) Maximal deformation with a \SI{200}{\gram} payload.}
    \label{fig:new_figure_7}
    \vspace{-0.8cm}
\end{figure}

The quantitative results of the parameter sweep are visualized in the contour plots of Fig.~\ref{fig:deformable_results}.These map the measured roundness ratio as calculated using equation~(\ref{equa_roundness}),  across the ($N, P$) parameter space. The color gradient visualizes the outcome of the roundness. In these trials, a grasp is defined as a failure if the gripper cannot lift the object, a slip if the object is lifted but drops during transport, and a success if it is stably transported to the target point. To reveal the physical mechanism behind this performance, the gripper's morphological change was also quantified (Fig.~\ref{fig:new_figure_7}). The experimental results reveals a distinct optimal status for each payload. For the \SI{100}{\gram} payload, the maximum roundness ratio $\approx 0.94$ is localized at $N \approx \SI{1.5}{\newton}$ and $P \approx \SI{100}{\kilo\pascal}$. As the payload increases to \SI{200}{\gram}, this optimal status shifts to $N \approx \SI{3.5}{\newton}$ and $P \approx \SI{50}{\kilo\pascal}$, achieving a best-case roundness ratio of $\approx 0.78$. Increasing the grasping force generally reduces the roundness ratio, leading to greater object deformation and a higher risk of damaging fragile or sensitive items. However, for low-mass objects (requiring lower gripping forces), increasing the inner pressure can help minimize deformation (maintaining a high roundness ratio), as shown in Fig.~\ref{fig:deformable_results} (a). Conversely, for heavier objects requiring larger grasping forces, increasing the inner pressure leads to more significant deformation (a lower roundness ratio), Fig.~\ref{fig:deformable_results} (b).

The experimental results indicate that balancing successful handling with minimal object deformation requires an optimal combination of force and provided pressure. Nonetheless, this study does not focus on feedback control; instead, it focuses on evaluating the roles of grasping force and inner pressure within the air pocket for achieving both successful and safe object handling.

\subsubsection{Grasping Different Objects Experiment Results}
\begin{figure}[htbp]
    \centering
    \includegraphics[width=\columnwidth]{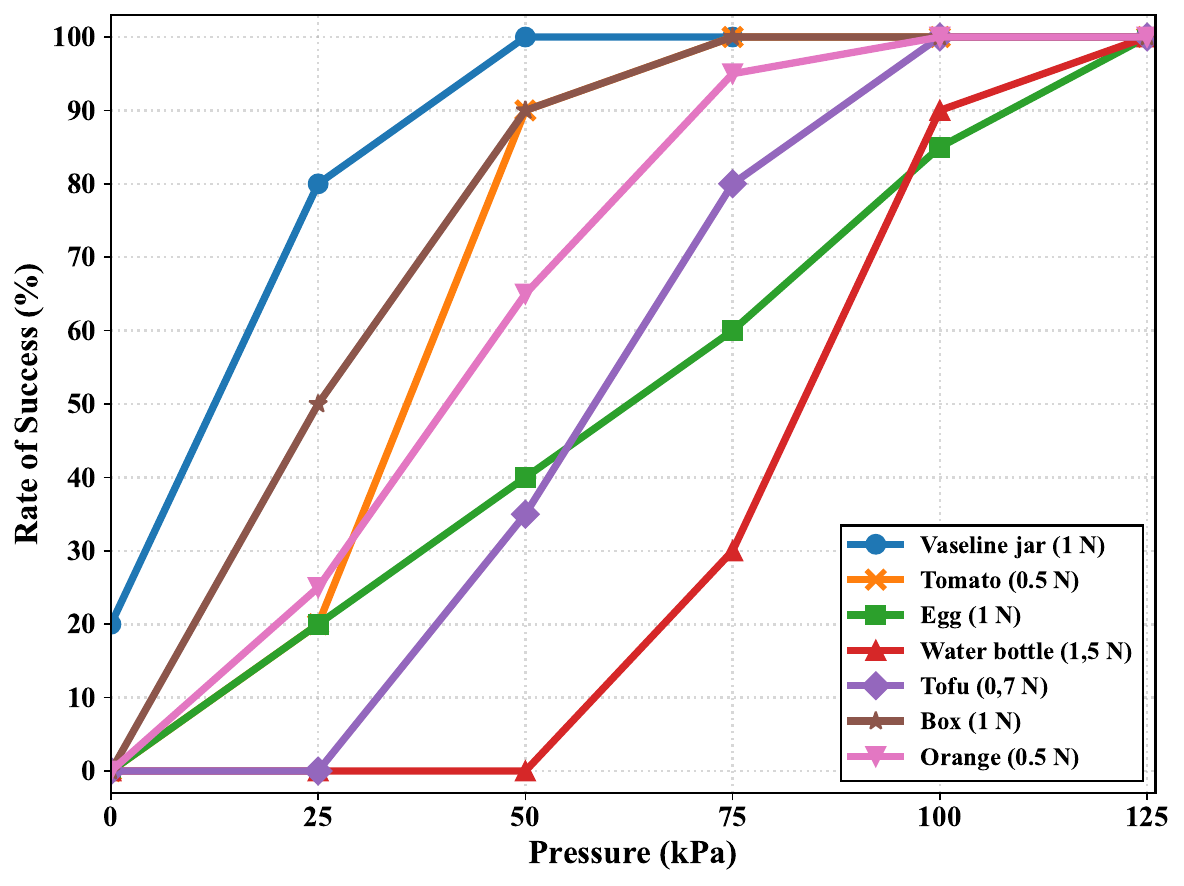}
    \caption{Grasping success rate as a function of actuation pressure for seven different objects, each tested at a pre-determined normal force.}
    \label{fig:versatility_plot}
    \vspace{-0.5cm}
\end{figure}

Each object was grasped 10 times, and the grasping success rate for each of the seven diverse objects is plotted as a function of actuation pressure in Fig.~\ref{fig:versatility_plot}. For all objects, the success rate demonstrates a strong positive correlation with the actuation pressure. At \SI{0}{\kilo\pascal}, the gripper can hardly handle the object. This confirmed that the small normal force is insufficient for a reliable grasp. As the pressure increases, the silicone pads inflate, increasing the contact area and friction, which leads to an increase in the success rate. For hard and heavy objects such as vaseline jars, boxes, or tomatoes, an inner pressure of 50 kPa is sufficient to achieve a success rate of over 90 \%. For elongated objects or those with rounded, smooth surfaces—such as water bottles and eggs—the inner pressure must be increased to 100 kPa to reach a success rate of over 80\%. For soft and fragile items such as tofu, an inner pressure of 75 kPa reaches an 80\% success rate. Finally, a 100\% success rate is achieved across all object types when the pressure is increased to 125 kPa.
The results show that the gripper's high adaptability to varying material properties and geometries. Rigid and slippery items generally achieved complete success at moderate pressures, while heavier payloads and more fragile objects required higher internal pressures to ensure stability. Moreover, the gripper effectively handled delicate and soft-surfaced items, such as tofu and tomato, by utilizing pressure-induced friction rather than relying on damagingly high normal forces. This demonstrates that using a suitable gripping force and inner pressure of air pockets, the hybrid mechanism can adapt to a wide range of shapes, surface textures and fragilities while maintaining object integrity.

%%%%%%%%%%%%%%%%%%%%%%%%%%%%%%%%%%%%%%%%%%%%%%%%%%%%%%%%%%%%%%%%%%%%%%%%%%%%%%%%
\section{Limitations \& Future Works}

The proposed hybrid gripper finger has demonstrated its effectiveness and safety in grasping various objects by adjusting the air pressure supplied to the soft component of the finger. However, the gripper finger still has several limitations. First, in this study, the soft pocket employs three bulges to regulate the gripper’s friction; however, the number and size of these bulges have not yet been optimized. In future work, we aim to optimize both the number and dimensions of the bulges to achieve higher gripping efficiency. Second, in the current study, we do not focus on the feedback control problem; instead, our primary objective is to investigate the fundamental correlation between friction modulation and the variable stiffness of the gripper's soft pocket. Implementing a dynamic feedback control system to determine the optimal normal force and internal pressure remains a subject for future work. This could be achieved by integrating tactile sensors to detect the contact force directly and employing reinforcement learning algorithms for fully autonomous grasping control. Finally, the current gripper design is relatively simple and thus somewhat bulky; future work will focus on optimizing the design to make it more compact and compatible with widely used robotic arms.

%%%%%%%%%%%%%%%%%%%%%%%%%%%%%%%%%%%%%%%%%%%%%%%%%%%%%%%%%%%%%%%%%%%%%%%%%%%%%%%%
\section{Conclusion}

In this study, a robotic gripper using the soft-rigid fingers that combine rigid shells with inflatable silicone pockets to modulate surface friction via air pressure was introduced. Experimental results show that increased pressure enhances friction, enabling stable grasping of heavy or slippery objects with minimal gripping force. Furthermore, the gripper safely handles fragile and deformable objects, demonstrating its versatility across diverse objects. The results show the potential of pressure-controlled friction for optimizing robotic interactions. Future work will focus on structural optimization and implementing intelligent control algorithms to automate the grasping process.

%%%%%%%%%%%%%%%%%%%%%%%%%%%%%%%%%%%%%%%%%%%%%%%%%%%%%%%%%%%%%%%%%%%%%%%%%%%%%%%%
% \newpage
\addtolength{\textheight}{-6cm}   % This command serves to balance the column lengths
% \newpage
\bibliographystyle{IEEEtran}
\balance
\bibliography{biblio}
% NOTE: You must add entries for Xiang2005 and Archard1957 to your biblio.bib file.
% For example:
% @article{Xiang2005,
%   title={Plane-strain bulge test for thin films},
%   author={Xiang, Y. and Chen, X. and Vlassak, J. J.},
%   journal={Journal of materials research},
%   volume={20},
%   number={9},
%   pages={2360--2370},
%   year={2005},
%   publisher={Cambridge University Press}
% }
% @article{Archard1957,
%   title={Elastic deformation and the laws of friction},
%   author={Archard, J. F.},
%   journal={Proceedings of the royal society of London. Series A. Mathematical and physical sciences},
%   volume={243},
%   number={1233},
%   pages={190--205},
%   year={1957},
%   publisher={The Royal Society}
% }

\end{document}